\pdfoutput=1
\documentclass[11pt]{article}

\usepackage[final]{acl}

\usepackage{times}
\usepackage{latexsym}
\usepackage{amsfonts}       % blackboard math symbols
\usepackage{amsmath}

\usepackage{multicol}
\usepackage{hyperref}       % hyperlinks
\usepackage{url}            % simple URL typesetting
\usepackage{booktabs}       % professional-quality tables
\usepackage{amsfonts}       % blackboard math symbols
\usepackage{nicefrac}       % compact symbols for 1/2, etc.
\usepackage{microtype}      % microtypography
\usepackage{xcolor}         % colors
\usepackage{float}
\usepackage[raster,most]{tcolorbox}
\usepackage{natbib}

\usepackage{booktabs}
\usepackage{multicol}
\usepackage{colortbl}
\newcommand{\E}{\mathbb{E}}
 \usepackage{multirow}
\usepackage[T1]{fontenc}
\usepackage[utf8]{inputenc}

\usepackage{microtype}

\usepackage{inconsolata}

\usepackage{graphicx}
\usepackage{tabularx}
\usepackage{enumitem}

\title{Toward Effective Reinforcement Learning Fine-Tuning for Medical VQA in Vision-Language Models}

\author{
\textbf{Wenhui Zhu\textsuperscript{1*}},
\textbf{Xuanzhao Dong\textsuperscript{1*}},
\textbf{Xin Li\textsuperscript{1*}},
\textbf{Peijie Qiu\textsuperscript{3*}},
\textbf{Xiwen Chen\textsuperscript{2*}},\\
\textbf{Abolfazl Razi\textsuperscript{2}},
\textbf{Aris Sotiras\textsuperscript{3}},
\textbf{Yi Su\textsuperscript{4}},
\textbf{Yalin Wang\textsuperscript{1}}\\
\textsuperscript{1}Arizona State University,
\textsuperscript{2}Clemson University,
\textsuperscript{3}Washington University in St.Louis\\
\textsuperscript{4}Banner Alzheimer’s Institute\\
\small{\textbf{Correspondence:} \href{mailto:email@domain.com}{wzhu59@asu.edu}}
}

\begin{document}
\maketitle
\begin{abstract}
% Recently, reinforcement learning (RL)-based tuning has shifted the trajectory of Multimodal Large Language Models (MLLMs), particularly following the introduction of Group Relative Policy Optimization (GRPO). However, directly applying it to medical tasks remains challenging for achieving clinically grounded model behavior. Motivated by the need to align model response with clinical expectations, we investigate three critical dimensions that affect the effectiveness of RL-based tuning in medical visual question answering (VQA): i) base model initialization strategy, ii) the role of medical semantic alignment and iii) the impact of length-based rewards on long-chain reasoning. We aim to investigate and uncover how to adapt RL-based tuning of MLLMs for medical applications. We further validate our analysis through systematic experiments on a large-scale medical VQA benchmark. Our investigation reveals that RL-based tuning outperforms the SFT in the medical MLLM domain.
Recently, reinforcement learning (RL)-based tuning has shifted the trajectory of Multimodal Large Language Models (MLLMs), particularly following the introduction of Group Relative Policy Optimization (GRPO). However, directly applying it to medical tasks remains challenging for achieving clinically grounded model behavior. Motivated by the need to align model response with clinical expectations, we investigate four critical dimensions that affect the effectiveness of RL-based tuning in medical visual question answering (VQA): base model initialization strategy, the role of medical semantic alignment, the impact of length-based rewards on long-chain reasoning, and the influence of bias. We conduct extensive experiments to analyze these factors for medical MLLMs, providing new insights into how models are domain-specifically fine-tuned. Additionally, our results also demonstrate that GRPO-based RL tuning consistently outperforms standard supervised fine-tuning (SFT) in both accuracy and reasoning quality.

% The code is available at \url{https://github.com/LLM-VLM-GSL/Discuss-RAG}.
\end{abstract}

\def\thefootnote{*}\footnotetext{These authors contributed equally to this paper.}

\section{Introduction}
%%%% say LLMs ability has enlarge LLM ability, go from SFT to Reinforcement learning, especially the GRPO
% Enabling large language models (LLMs) with reasoning abilities has become a prominent focus in contemporary research, as it represents a crucial step toward achieving Artificial General Intelligence (AGI). A common approach to addressing this challenge is supervised fine-tuning (SFT), which utilizes structured datasets containing explicit intermediate reasoning steps to guide LLMs toward human-like responses. This tuning strategy has demonstrated substantial performance gains. For example, the OpenAI-O1 model was among the first to show strong reasoning ability after being trained on high-quality chain-of-thought (CoT) data. However, SFT has notable limitations: it requires significant computational resources and involves considerable manual effort in curating both reasoning paths and dataset content.

Encouraged by its success in Large Language Models (LLMs), researchers have extended Group Relative Policy Optimization (GRPO)~\cite{shao2024deepseekmath} to multimodal domains such as image understanding, audio processing, UI design, medical analysis, and physical world interaction~\cite{chen2025r1v,shen2025vlm,lai2025med,lu2025ui,xia2025gui,zhao2025r1,azzolini2025cosmos,zhu2025retinalgpt,chen2025dra}. Specifically, studies like~\cite{huang2025vision,zhou2025r1zerosahamomentvisual} report promising results, including emergent reasoning in compact models. However, applying GRPO-based RL to medical visual question answering (VQA), which demands clinically accurate outputs, remains underexplored.

In this work, we analyze GRPO-based RL for medical Multimodal Large Language Model (MLLMs) across five critical aspects:

\noindent\textbf{(1) Training from Scratch vs. Fine-Tuning.}  
Training from scratch allows for unconstrained reasoning exploration, but initializing from an instructionally fine-tuned model stabilizes training and accelerates convergence~\cite{zhang2023instruction,chung2024scaling}. To investigate this trade-off, we conduct experiments comparing both approaches. Our results show that prior instruction supervised fine-tuning (SFT) improves both answer accuracy and domain alignment.

\noindent\textbf{(2) Rewarding Medical Semantic Accuracy.}  
Generic rewards for format or output correctness are insufficient for clinical tasks. We introduce a medical semantic reward using LLM-generated evaluations, guided by prompt engineering. This significantly boosts both response quality and clinical alignment.

\noindent\textbf{(3) Does Longer Reasoning alone Help?}  
% Although deeper reasoning can improve accuracy, we found that applying length-based rewards (Extended Chain Reward (ECR) and Correctness-Weighted Length Reward (CWR)) often leads to verbose and less accurate answers. To address this, we incorporate Dr.GRPO, which improves performance by reducing length bias. Our results show that encouraging longer outputs alone is ineffective for medical VQA.
Although many studies have shown that deep reasoning can be beneficial~\cite{cheng2024vision, kumar2025llm}, we found that relying solely on length-based rewards (i.e., Extended Chain Reward (ECR) and Correctness-Weighted Length Reward (CWR)) often leads to verbose and less accurate answers. This observation calls into question the effectiveness of using length-based rewards along as a mechanism to promote meaningful long-form reasoning in medical VQA.

\noindent\textbf{(4) Does bias exist in medical MLLMs?} Normalization is commonly used to stabilize training. However, recent study~\cite{liu2025understanding} suggest that the question-level normalization may inadvertently bias model behavior, encouraging the generation of longer but incorrect responses by amplifying the per-token gradient signal. To further evaluate its impact in medical VQA, we implement Dr.GRPO~\cite{liu2025understanding}. Our results demonstrate its effectiveness in improving both answer accuracy and token efficiency.

\noindent\textbf{(5) SFT vs. GRPO-based RL tuning.} SFT is a widely adopted strategy to enhance the reasoning ability of MLLMs. To compare its effectiveness against GRPO-based RL tuning, we implemented three different SFT approachs and additionally evaluated two publicly available MLLMs. Our results show that the GRPO-based RL tuning consistently outperforms SFT methods, yielding higher answer accuracy and more clinically grounded responses.

Our main contributions can be summarized as follows: 
\begin{itemize}[itemsep=-0.2em, topsep=0em]
    \item We present a systematic analysis of GRPO-based RL in medical MLLMs, focusing on initialization strategies, medical semantic alignment, the impact of length-based rewards and bias-related behaviors.
    \item We validate our findings through large-scale experiments on medical VQA benchmarks, offering practical insights into aligning RL with clinically meaningful behavior.
    \item We find that GRPO-based RL tuning outperforms traditional fine-tuning methods (e.g.,SFT), highlighting its potential for developing more capable and aligned medical MLLMs.
\end{itemize}

\section{Preliminary}\label{Sec:pre}
Instead of relying on an extra reward and value model as in PPO~\cite{schulman2017proximalpolicyoptimizationalgorithms}, GRPO simplifies the process by using the average reward from the policy model's sampled responses as a baseline for advantage estimation. Specifically, given the an input question $q$, we first sample a group of responses $\{o_1, o_2, \cdots, o_G\}$ from the policy model $\pi_{\theta_{\text{old}}}$ and compute corresponding rewards $\mathbf{r} :=\{r_1, r_2, \cdots, r_G\}$. Then GRPO seeks to optimize the following objective and update the model $\pi_\theta$, denoted as:
{\footnotesize
\begin{equation}
\begin{split}
    \mathcal{J}_{GRPO}&(\theta) := \E_{q\sim p_Q, \{o_i\}_{i=1}^G \sim \pi_{\theta_{\text{old}}}(O|q)} \\
    &  \Bigg\{\frac{1}{G}\sum_{i=1}^G   \min \Bigg[ \frac{\pi_\theta(o_i |q)}{\pi_{\theta_{\text{old}}}(o_i |q)} \hat{A}_{i},  \\
    &\text{clip} \left( \frac{\pi_\theta(o_{i} | q)}{\pi_{\theta_{\text{old}}}(o_{i} | q)}, 1 - \epsilon, 1 + \epsilon \right)  \hat{A}_{i} \bigg] \\
    &- \beta \cdot D_{KL}(\pi_\theta||\pi_{\text{ref}}) \Bigg\},
\end{split}
\label{eq:GRPO-obj}
\end{equation}
}
where the advantage is denoted as:
\begin{equation}\label{eq:advantage}
    \hat{A}_{i} := \frac{r_i- {\rm mean}(\mathbf{r})}{{\rm std}(\mathbf{r})}
\end{equation}
Here, $\epsilon$ and $\beta$ denote the PPO clipping hyperparameter and weight of KL-divergence regularization, respectively. We omit the token-level average per response in Eq.~\ref{eq:GRPO-obj} for simplicity. Additionally, we strictly follow DeepSeek-R1~\cite{guo2025deepseek}, using rule-based reward (i.e., format and response reward) as our base reward design.

\section{Emperical Studies on RL tuning}\label{Sec:method}
We selected Qwen2-VL-2B~\cite{wang2024qwen2vlenhancingvisionlanguagemodels} as our base model. All experiments were conducted on the subset of the PMC-VQA~\cite{zhang2023pmc} benchmark, including 10K training samples and 7K testing samples. In this section, we analyze and answer four key questions that influence the effectiveness of GRPO-based RL tuning in medical VQA. The training parameters follow the settings used in previous work~\cite{visualthink}, and additional details are provided in Appendix~\ref{appendix:ID}. Here, response accuracy, similarity score, perplexity~\cite{chung2025qwen}, thinking reward~\cite{jiang2023mistral}, and thinking token length are considered as metrics wherever such measurements are available. We outline the details of metrics in Appendix~\ref{appendix:eval}.
\subsection{From Scratch vs. Fine-Tuning.} Recent studies on GRPO-based RL tuning in MLLMs have typically relied on base models that were already instruction fine-tuned~\cite{chen2025r1v, zheng2025easyr1, shen2025vlmr1, wang-2025-open-r1-video}. However, these models often fail to exhibit the "aha moment" in their learning curves, suggesting that instruction tuning may hinder the reasoning exploration. Indeed, ~\cite{visualthink} shows that cold-start GRPO-based RL without SFT can more effectively promote reasoning behavior in the MLLMs setting.
To examine the role of domain knowledge, we compare GRPO-based RL on Qwen2-VL-2B (trained from scratch) and Qwen2-VL-2B-Instruct (instruction-tuned). As shown in Tab.~\ref{tab:main_result}, the scratch-trained model has a higher similarity score and a +1.61 gain in thinking reward, meaning its reasoning is more aligned and useful.
However, it also shows lower accuracy and higher perplexity Score, which suggests less correct and less fluent answers. 

This means that while training from scratch encourages more reasoning, it lacks the medical knowledge and language fluency that instruction tuning provides (see Fig.~\ref{fig:scratchRL} in Appendix~\ref{appendix:vis} for examples).
These results show that how a model is initialized affects its performance. Instruction-tuned models give more accurate and fluent answers. For medical VQA, utilizing domain-specific pretraining methods (e.g., cold-start, pretraining) helps strike a balance between answer quality and reasoning.

\subsection{Medical Semantic Alignment.}\label{Sec:semantic-alignment}
Aligning the model’s reasoning path with the target task can enhance the effectiveness of GRPO-based RL tuning. To further investigate the impact of medical semantic alignment in the medical VQA setting, we introduce a semantic alignment reward that encourages model responses to match the judgments of predefined expert LLMs. Specifically, we use Qwen2-VL-2B-Instruct as the base model and employ BioGPT~\cite{luo2022biogpt} and BioMistral~\cite{labrak2024biomistral} as the reference LLMs. 

As illustrated in Fig.~\ref{fig:template}, we design a prompt template in which the reference LLM is asked to assess whether the reasoning enclosed within the \texttt{<think></think>} tags is clinically grounded during training. If the reasoning is valid, the LLM responds with "Yes," and a reward of 1 is assigned; otherwise, it responds with "No," and a reward of 0 is given. Results in Tab.~\ref{tab:main_result} show that adding semantic alignment improves both performance and reasoning quality. Accuracy increases by 1.82\%, and the Similarity Score improves by 0.25, indicating that the model’s reasoning becomes more semantically aligned with reference answers. Overall, the integration of medical semantic rewards leads to notable improvements in both accuracy and reasoning depth. Example visualizations are provided in Fig.\ref{fig:med_align} in Appendix~\ref{appendix:vis}.

% As shown in Tab.\ref{tab:main_result}, the incorporation of medical semantic rewards leads to clear accuracy and reasoning improvement. Example visualizations are provided in Fig.~\ref{fig:med_align} in Appendix.~\ref{appendix:vis}.
% Example visualizations are provided in the Appendix~\ref{appendix:vis}.

% \begin{figure}
%     \centering
%     \includegraphics[width=1\columnwidth]{image/medical-vllm-grpo.pdf}
%     \caption{Illustration of the horizontal bar chart \textbf{(A)} depicting response accuracy (highlighted in green) and the dotted curves \textbf{(B)} representing thinking token length (highlighted in blue). The term \textbf{Med-S-Align} denotes medical semantic alignment. The base model variants (i.e., Qwen-VL-2B and Qwen-VL-2B-Instruct) are highlighted in brown. Additional details are provided in Sec.~\ref{Sec:method}.}
%     \label{fig:accuracylength}
% \end{figure}

\begin{figure}
    \begin{tcolorbox}[title=Template for Semantic LLM Judgment, label=meaningless_response]
    \textbf{Prompt:} Evaluate the following medical statement for semantic correctness and clinical coherence: 
    
    Given Statement: Reasoning Text (e.g., text between <think></think>)  
    
    Answer 'Yes' if the statement is medically coherent and accurate, or 'No' otherwise. 
    
    Answer:
    % \textbf{Reward:} The reward will return 1 if the answer is 'Yes', otherwise return 0. 
\end{tcolorbox}
    \caption{Illustration of the prompt template used to evaluate the effectiveness of medical semantic alignment. See more details in Sec.~\ref{Sec:semantic-alignment}. }
    \label{fig:template}
    \vspace{-0.5cm}
\end{figure}

\begin{table*}[!t]
\caption{An illustration of all our experimental results. Specifically, \textbf{Qwen2-VL-2B-Instruct}, \textbf{LLaVA-7B-v1.5} and \textbf{Llama-3.2-11B-Vision-Instruct} are used directly without additional fine-tuning. For experiments involving fine-tuning, \textbf{Qwen2-VL-2B-Instruct} is used as the default base model unless otherwise specified (e.g., training from scratch with \textbf{Qwen2-VL-2B}). Additional experimental details are provided in Sec.\ref{Sec:method}. }
\vspace{-0.1in}
\label{tab:main_result}
\resizebox{\textwidth}{!}{%
\begin{tabular}{lccccccc} \toprule
\multicolumn{1}{c}{\textbf{Model}} &\textbf{Fine-Tuning Data Scale} &\textbf{Accuracy }$\uparrow$ & \textbf{Similarity Scores}$\uparrow$ & \textbf{Perplexity Scores}$\downarrow$ & \textbf{Thinking Rewards}$\uparrow$ & \textbf{Thinking Token Length} \\ 
\midrule
Qwen2-VL-2B-Instruct & - & 47.29 & - & - & - & - \\
\multicolumn{1}{@{}l}{\bfseries \textit{SFT-based Training}} \\
\hspace{1em}--- LoRA & 10K & 45.98 & - & - & - & - \\
\hspace{1em}--- Full Fine-Tuning & 10K & 52.00 & - & - & - & - \\
\hspace{1em}--- DPO Fine-Tuning & 10K & 46.97 & - & - & - & - \\
\multicolumn{1}{@{}l}{\bfseries \textit{Training from Scratch}} \\
Qwen2-VL-2B &  &  &  &  &  &  \\
\hspace{1em}--- GRPO & 10K & 51.56 & 0.49 ($\pm 0.19$) & 14.16 ($\pm 2.80$) & 9.27 ($\pm 2.09$) & 141.46 ($\pm 69.10$) \\
\multicolumn{1}{@{}l}{\bfseries \textit{RL-based Training}} \\
Qwen2-VL-2B-Instruct &  &  &  & &  &  \\
\hspace{1em}--- GRPO & 10K & 58.04 & 0.21 ($\pm 0.24$) & 13.28 ($\pm 19.64$) & 7.66 ($\pm 2.49$) & 66.41 ($\pm 74.34$) \\
\hspace{1em}--- GRPO + Semantic Alignment & 10K & 59.86 & 0.46 ($\pm 0.19$) & 36.54 ($\pm 10.97$) & 8.07 ($\pm 2.63$) & 64.42 ($\pm 24.02$) \\
\hspace{1em}--- GRPO + Semantic Alignment + ECR & 10K & 50.17 & 0.65 ($\pm 0.07$) & 20.54 ($\pm 2.92$) & 9.42 ($\pm 1.91$) & 338.14 ($\pm 95.69$) \\
\hspace{1em}--- GRPO + Semantic Alignment + CWR & 10K & 54.68 & 0.61 ($\pm 0.11$) & 18.45 ($\pm 2.40$) & 7.82 ($\pm 3.52$) & 224.06 ($\pm 73.72$) \\
\hspace{1em}--- Dr.GRPO & 10K & 61.09 & 0.24 ($\pm 0.25$) & 11.10 ($\pm 12.63$) & 4.17 ($\pm 3.76$) & 76.11 ($\pm 80.27$) \\
\midrule
\multicolumn{1}{@{}l}{\bfseries \textit{Other Models}} \\
LLaVA-7B-v1.5 & - & 11.8 & 0.19 ($\pm 0.24$)* & - & - & - \\
Llama-3.2-11B-Vision-Instruct & - & 22.92 & 0.33 ($\pm 0.27$)* & - & - & - \\
\bottomrule
\end{tabular}%
}
\end{table*}

\subsection{The Influence of Long-Chain Reasoning.}
Chain-of-thought (CoT) reasoning has been shown to improve performance in large language models~\cite{team2025kimi,guo2025deepseek,llmsdemystifying}, but its role in medical VQA is less understood~\cite{visualthink,zhang2024improve,dong2024insight}. To explore this, we incorporated an Extended Chain Reward (ECR) during GRPO-based RL tuning of Qwen2-VL-2B-Instruct, alongside a medical semantic alignment reward. ECR incentivizes longer reasoning chains by assigning an additional reward based on output length.

As shown in Tab.~\ref{tab:main_result}, adding ECR increases token Length by 273.72 and improves Similarity Score by 0.19 and thinking reward by 1.35. However, these gains come at the expense of a 7.87\% drop in accuracy, indicating that the model begins to favor more verbose and elaborate reasoning at the cost of factual correctness. To mitigate this, we introduced a Correctness-Weighted Length Reward (CWR), which incentivizes long responses only when the final answer is correct. This is achieved through symbolic and string-based correctness checks. As shown in Tab.~\ref{tab:main_result}, when combined with semantic alignment, CWR enhances fluency and structure coherence, as evidenced by improvements in perplexity and similarity Score. However, it remains 3.36\% lower than the baseline GRPO model. Although token length still increases considerably, the model continues to exhibit tendencies toward exploiting length-based incentives.
These findings suggest that relying on length-focused reward can lead to verbosity. Striking an appropriate balance between factual accuracy and high-quality reasoning remains a key challenge in medical VQA. Additional illustrative examples are provided in Appendix~\ref{appendix:vis}, Fig.~\ref{fig:longc}.

\subsection{Unbiased GRPO.}
% In previous tuning experiments, applying vanilla GRPO combined with length-based rewards yielded overly verbose outputs, compromising factual accuracy. This behavior aligns with recent findings about the advantage estimation in GRPO (i.e., Eq.~\ref{eq:advantage}). Specifically, its standard deviation normalization can introduce bias by encouraging long or extreme responses regardless of correctness.

% In GRPO-based RL tuning, the length of responses per question varies. This misalignment against the pretraining stage (i.e., all tokens are packed into context with fixed length) may lead to unintentionally model behavior in GRPO-based RL tuning. Specifically, the token-level normalization leads model to generate 
% Normalization may unintentionally distort model behaviors. Specifically, the token-level normalization can reduce the impact of negative advantage, encouraging model to generate longer yet incorrect responses. Moreover, the question-level normalization may further exacerbate this misalignment, particularly in the presence of overconfident questions (i.e., questions are either too simply or too difficult). Collectively, these effects raise concerns about the reliability of reasoning behaviors in medical VQA.

% To investigate the influence of normalization, we adopt Dr. GRPO~\cite{liu2025understanding}, which removes both standard deviation normalization and token-level averaging. Instead, the advantage is computed as a simple difference from the group mean reward:
Normalization can distort model behavior. Token-level normalization may weaken the effect of negative advantages, leading the model to generate longer but incorrect answers. Question-level normalization can exacerbate this issue, particularly with overconfident questions (i.e., those that are too easy or too hard). These effects raise concerns about the reliability of reasoning in medical VQA.
To investigate this, we apply Dr. GRPO~\cite{liu2025understanding}, which removes standard deviation normalization and token-level averaging. It computes advantage as a simple difference from the group mean reward.

% \begin{equation}
%     \hat{A}_{i, t} = \widetilde{r}_i = r_i- {\rm mean}(\mathbf{r})
% \end{equation}
\begin{equation}
    \hat{A}_{i}  := r_i- {\rm mean}(\mathbf{r})
\end{equation}
This formulation provides a more stable and interpretable reward signal by reducing the influence of response length and preventing overly sharp gradients. We integrate Dr.GRPO into the GRPO-based RL tuning of Qwen2-VL-2B-Instruct, using the same training setup.
As shown in Tab.~\ref{tab:main_result}, Dr.GRPO achieves the highest accuracy among all configurations, surpassing the standard GRPO baseline by 3.05\%. It also yields improvements in both Perplexity and Similarity Score,  indicating enhanced fluency and semantic alignment. These results suggest that removing normalization mechanisms facilitates more stable training dynamics and improves the alignment between reasoning quality and answer correctness. Overall, Dr. GRPO provides a more reliable and interpretable optimization signal, supporting the generation of clinically relevant and efficient responses in medical VQA.
% (as illustrated in Fig.~\ref{fig:longc}). 
% Example visualizations are provided in the Appendix~\ref{appendix:vis}.

\section{SFT vs. GRPO-based RL tuning}\label{Sec:sft-vs-rl}
Given that SFT is another widely used approach to endow MLLMs with reasoning ability, an important question arises in the context of clinically meaningful RL-tuned MLLMs: \textit{Which performs better in medical VQA — GRPO-based RL or SFT-based training?} 
To explore this comparison, we follow prior SFT work~\cite{Qwen2-VL-Finetuning}, evaluating three SFT strategies (i.e., full fine-tuning, LoRA~\cite{hu2022lora} and DPO fine-tuning~\cite{rafailov2023direct}) against the GRPO-based RL tuning, along with the aforementioned modification. In all above experiments, Qwen2-VL-Instruct serves as the base model. In addition to our fine-tuned models, we include two publicly available MLLMs: LLaVA-7B-v1.5~\cite{liu2023visual} and Llama-3.2-11B-Vision-Instrct~\cite{grattafiori2024llama}, both of which are already trained using supervised fine-tuning. 

In a medical VQA setting, GRPO-based RL tuning consistently outperforms SFT-based approaches. As shown in Tab.~\ref{tab:main_result}, the two public MLLMs exhibit a clear gap in accuracy and semantic alignment compared to our tuned models. Among the SFT methods, full fine-tuning achieves the highest accuracy of 52.00. However, applying GRPO-based RL tuning to the same base model results in a significant improvement in performance. These results suggest that while SFT enables the model to imitate reasoning patterns observed in data, it remains limited in its capacity to induce genuine reasoning ability. Moreover, we also observe that models lose CoT reasoning ability after undergoing SFT (examples refer to Appendix~\ref{appendix:vis}, Fig.~\ref{fig:SFT-RL}).
In contrast, GRPO-based RL training allows the MLLMs to automatically explore and exploit medically meaningful reasoning behavior.

\section{Conclusion}
In this work, we investigate the gap between GRPO-based RL tuning and clinically grounded MLLMs. We first examine the effectiveness of instructional fine-tuning, medical semantic alignment, and unbiased GRPO in improving answer accuracy, while also critically assessing the limitations of length-based rewards in promoting long-chain reasoning. Additionally, our findings show that GRPO-based RL consistently enhances the reasoning ability of medical MLLMs. We believe that this study offers valuable insights for advancing the development of clinically meaningful MLLMs and can inform future research in the medical AI community.

\newpage
% \section*{Limitation}
% In this work, we investigate the gap between GRPO-based RL and clinically grounded medical VLMs. While we conduct a multidimensional analysis supported by comprehensive experiments, we acknowledge two primary limitations of our study. \textbf{(1).} Due to limited computational resources, we only evaluated a subset of the PMC-VQA benchmark. Experiments on the full dataset may yield broader and more generalizable insights. \textbf{(2).} The development of GRPO-based RL methods is advancing rapidly. We believe that incorporating emerging techniques~\cite{chen2025dra} could further enhance the clinical reasoning capabilities of medical VLMs.
\section{Limitations}

Despite the promising results of applying GRPO-based reinforcement learning to medical MLLMs, our study has several limitations that open avenues for future research:

\begin{enumerate}
    \item \textbf{Scalability to Larger Datasets:} Our experiments are conducted on a subset of the medical dataset (PMC-VQA). While this setting allows for focused evaluation, it may limit the generalizability of our findings. Future work should consider expanding the study to larger and more diverse medical datasets, which may expose new challenges in model robustness and alignment with real-world clinical variability.

    \item \textbf{Model Size and Capacity:} All experiments in this work are conducted using Qwen2-VL-2B, a relatively small multimodal model. While this choice ensures training efficiency and interpretability of RL dynamics, it may not fully reflect the behavior of larger foundation models. Extending GRPO-based tuning to larger-scale MLLMs (e.g., $>$7B parameters) could provide insights into the scalability and generalization capabilities of our empirical study.

    \item \textbf{Incorporation of Expert-Labeled Chain-of-Thought (CoT) Data:} Currently, our approach does not leverage any expert-labeled reasoning traces or CoT annotations dataset. An interesting direction for future work is to explore the integration of CoT data, either through pretraining or cold-start initialization, to enhance models’ reasoning capabilities before RL tuning. This could help bridge the gap between language alignment and step-wise clinical logic.

    \item \textbf{Reasoning Limitations SFT:} As noted in our findings, models fine-tuned via SFT alone often fail to acquire robust reasoning abilities, especially in complex medical scenarios. This highlights a fundamental limitation of supervised fine-tuning when reasoning is not explicitly annotated. Investigating alternative strategies—such as integrating reasoning-aware objectives or hybrid SFT-RL pipelines—may offer promising solutions to enhance multi-step inference in medical MLLMs.
\end{enumerate}

\bibliography{main}

% \appendix

\clearpage
\appendix

\renewcommand{\thefigure}{S\arabic{figure}}
\renewcommand{\thetable}{S\arabic{table}}

\section{Implementation Details}\label{appendix:ID}
For distributed training, we employed DeepSpeed with ZeRO Stage 2 and \texttt{bfloat16} mixed precision. All experiments were conducted on a single machine equipped with four NVIDIA A100 GPUs, each with 80GB of memory. We launched four training processes using the standard DeepSpeed multinode launcher and disabled offloading for both optimizer and model parameters. The environment was configured for local execution with static rendezvous, no CPU fallback, and no TPU usage. The main training function was set to \texttt{main}, and communication was established via port 44326.

We used either the \textsf{Qwen2-VL-2B} or \textsf{Qwen2-VL-2B-Instruct} model as the backbone, trained on a multimodal subset of the PMC-VQA dataset. The input resolution was limited to 401{\texttimes}408 pixels, with a maximum prompt length of 1024 tokens. Training was conducted for two epochs using a per-device batch size of 1 and gradient accumulation over 2 steps. Mixed-precision training with \mbox{\texttt{bfloat16}} was enabled, while gradient checkpointing was disabled. Flash Attention 2 was used for efficient attention computation. Logging was performed at every step, and checkpoints were saved every 100 steps.

The model was trained for a total of 1500 steps with a learning rate of $1 \times 10^{-6}$ and a temperature of 1.0. To facilitate Guided Response Preference Optimization (GRPO), we set the maximum response length to 700 tokens. At each optimization step, 8 responses were sampled, and a KL divergence coefficient of 0.04 was applied to regularize training.

\begin{table}[h]
    \centering
    \caption{\textbf{Hyper-parameters}}
    \resizebox{1\columnwidth}{!}{
    \begin{tabular}{l c}
        \hline
        \toprule
        \textbf{Setting} & \textbf{Value} \\
        \hline
        Batch Size per Device & 1 \\
        Gradient Accumulation Steps & 2 \\ 
        Training Steps & 1500 \\
        Learning Rate & $1 \times 10^{-6}$ \\
        Temperature & 1.0 \\
        Maximum Response Length & 700 \\
        Number of Responses per GRPO Step & 8 \\
        KL Coefficient & 0.04 \\
        \hline
    \end{tabular}}
    \label{tab:exp_settings}
\end{table}

\section{Evaluation Metrics}\label{appendix:eval}
To comprehensively assess the quality of the model’s intermediate reasoning, we employ a diverse set of metrics, including similarity score, perplexity, thinking reward, and reasoning token length.

\subsection{Similarity Score}
Similarity Score measures how well the model’s reasoning aligns with the reference answer. We compute the semantic similarity between the reasoning and the ground-truth answer using a pretrained cross-encoder model (cross-encoder/stsb-roberta-base). This captures whether the model’s internal reasoning is semantically consistent with the correct final answer. 

\subsection{Perplexity Score}
Perplexity evaluates the fluency and linguistic quality of the model’s reasoning. We compute perplexity over the reasoning using a pretrained biomedical language model (microsoft/biogpt), following a standard left-to-right likelihood estimation. This metric captures how coherent and well-formed the reasoning appears from a language modeling perspective. Lower perplexity indicates more fluent, consistent, and syntactically stable reasoning.

\subsection{Thinking Reward}
Thinking Reward assesses the usefulness and relevance of the model’s reasoning content. To compute this score, we prompt a pretrained language model (mistral-7B-instruct) with the question, reference answer, and the reasoning generated from model, and ask it to assign a score from 1 to 10 based on the quality of the reasoning. This provides an external evaluation of whether the reasoning meaningfully contributes to answering the question.

\subsection{Thinking Token Length}
Thinking Token Length quantifies the length of the model’s reasoning by computing the total number of tokens. This metric provides a basic measure of reasoning verbosity and helps analyze the relationship between reasoning length and quality.

\section{Visualization}\label{appendix:vis}

\begin{figure*}
    \includegraphics[width=1\textwidth]{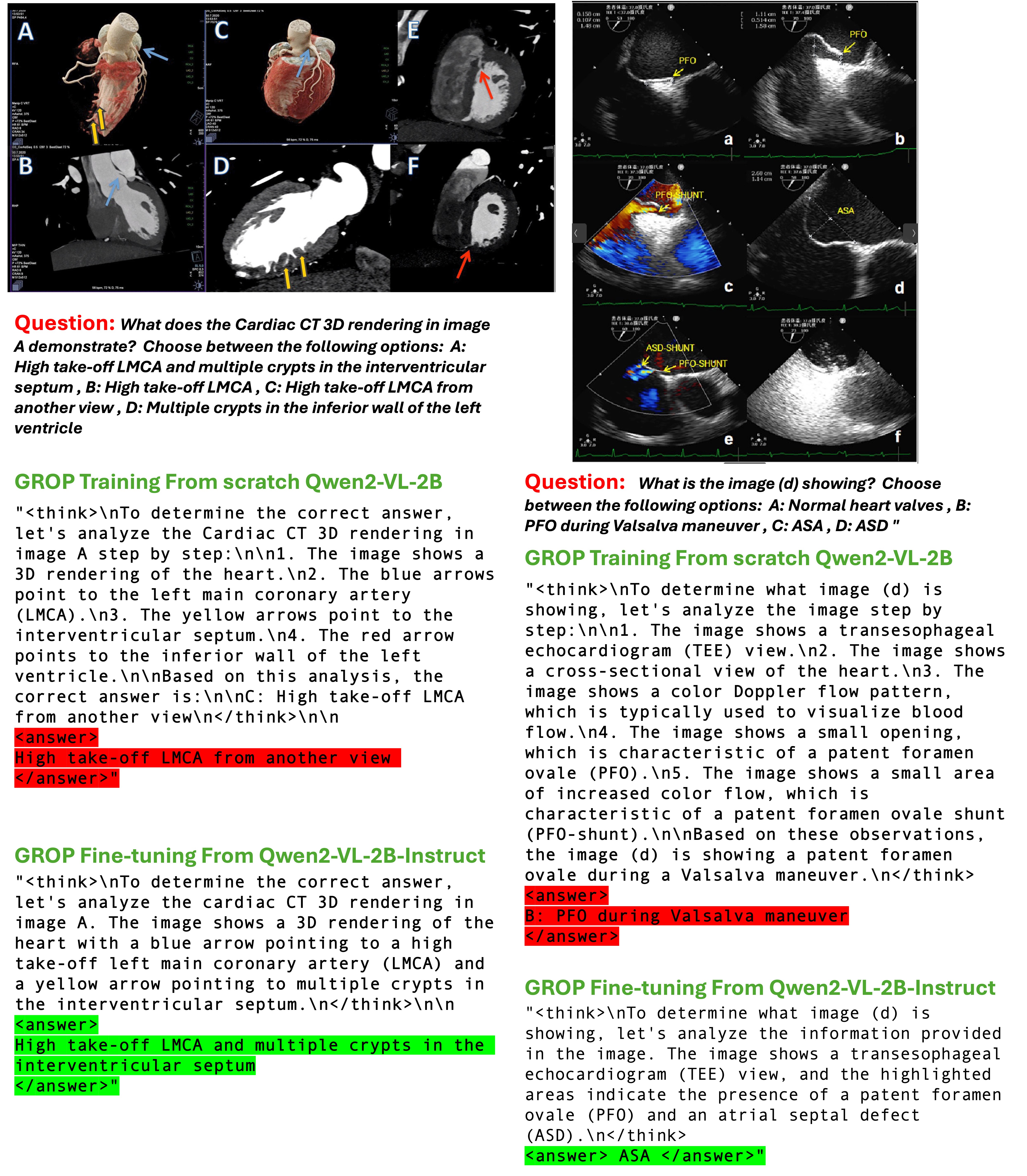}
    % \vspace{-0.1cm}
    \caption{Visual comparison of reasoning outputs on two medical imaging questions. Red highlights indicate incorrect answers, while green highlights indicate correct answers. Although training Qwen2-VL-2B from scratch with GRPO-based RL tuning model generates longer  sequences, its reasoning is often redundant and inaccurate. GRPO-based RL tuning based on Qwen2-VL-2B-Instruct produces more concise and clinically accurate reasoning, leading to correct answers. }
    \label{fig:scratchRL}
\end{figure*}

\begin{figure*}
    \includegraphics[width=1\textwidth]{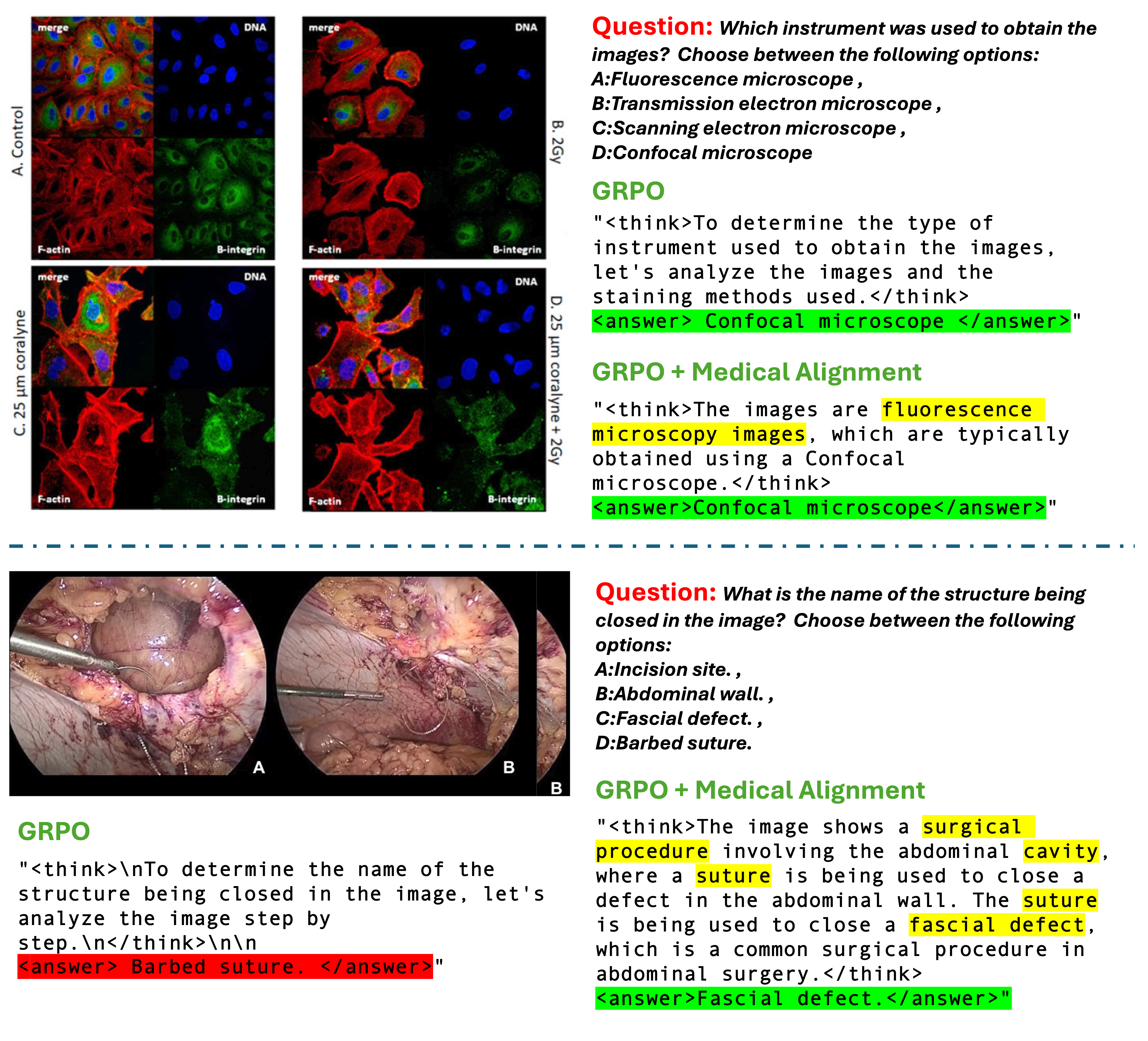}
    % \vspace{-0.1cm}
    \caption{Examples showing that medical alignment improves visual reasoning. Correct answers are shown in green, incorrect in red, and medical knowledge is highlighted in yellow. With medical alignment, the model produces more accurate and informed responses by grounding its reasoning in domain-specific knowledge. }
    \label{fig:med_align}
\end{figure*}

\begin{figure*}
    \includegraphics[width=1\textwidth]{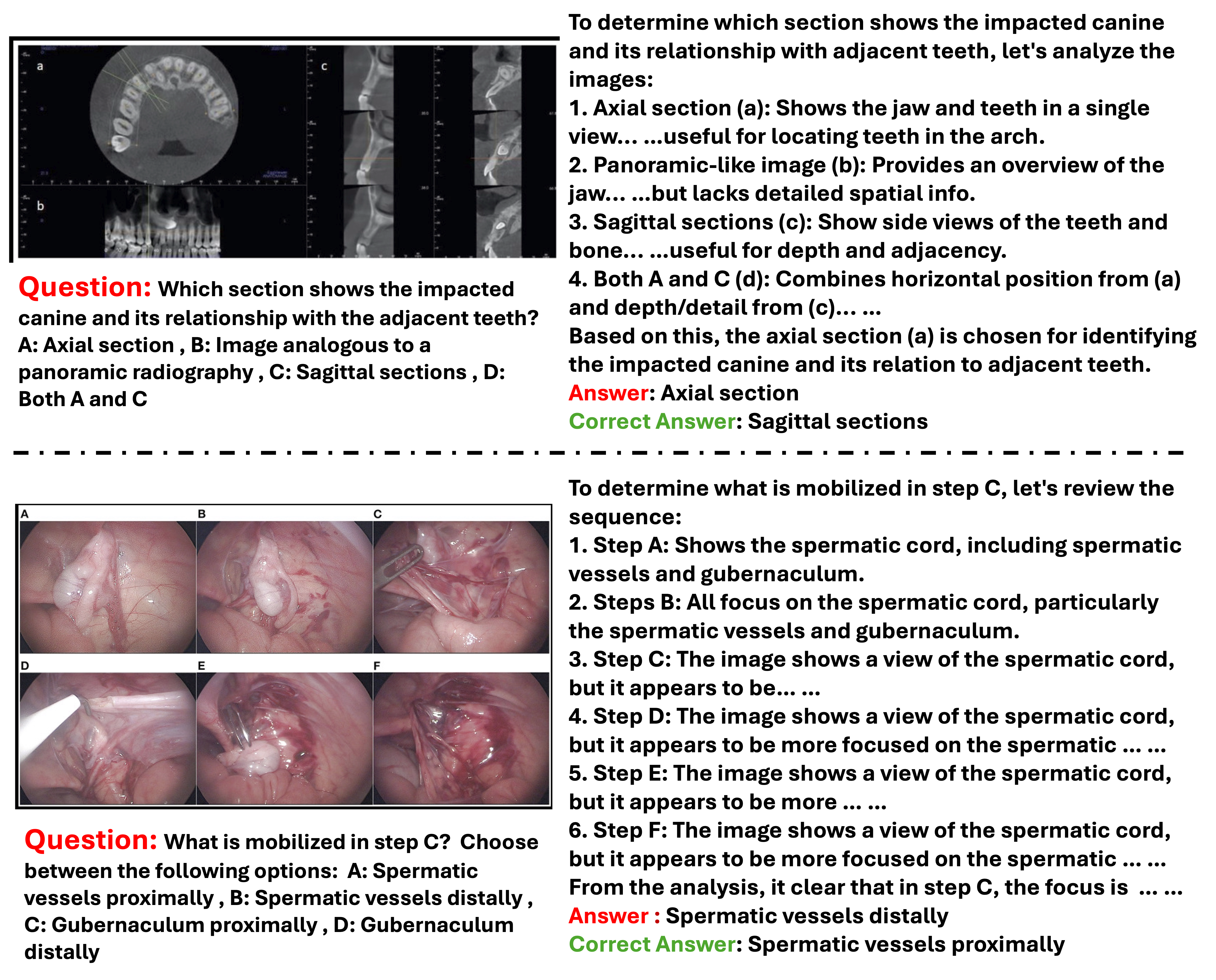}
    % \vspace{-0.1cm}
    \caption{Examples of incorrect but verbose reasoning in long-chain answers. Although the model generates extensive intermediate thinking steps, the reasoning is often repetitive, includes irrelevant details, and ultimately leads to an incorrect answer. }
    \label{fig:longc}
\end{figure*}

\begin{figure*}
    \includegraphics[width=1\textwidth]{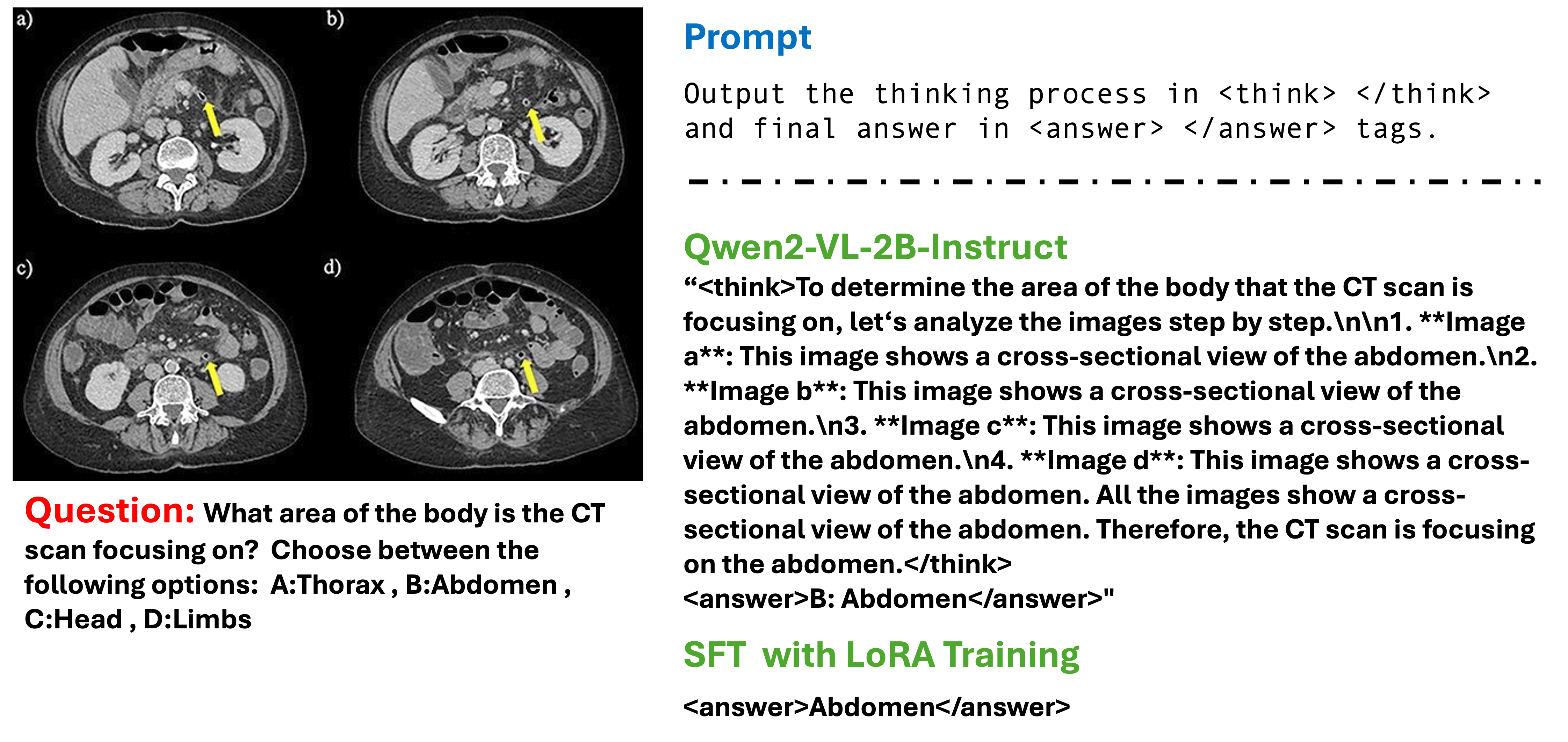}
    % \vspace{-0.1cm}
    \caption{ Comparison between the original Qwen2-VL-2B-Instruct and its LoRA fine-tuned variant. While the original model generates step-by-step visual reasoning to support its prediction, the LoRA-SFT version directly outputs the answer without any intermediate explanation.}
    \label{fig:SFT-RL}
\end{figure*}

% \begin{figure}[h]
%     \includegraphics[width=0.5\textwidth]{image/fig_longc.png}
%     \caption{\textbf{Examples of incorrect predictions in long-chain reasoning. Despite the length and apparent thoroughness of the reasoning, the model ultimately selects incorrect answers. }}
%     \label{fig:longc}
% \end{figure}
% This is an appendix.

\end{document}